\documentclass[conference]{IEEEtran}
\IEEEoverridecommandlockouts
\usepackage{cite}
\usepackage{amsmath,amssymb,amsfonts}
\usepackage{algorithmic}
\usepackage{graphicx}
\usepackage{textcomp}
\usepackage{xcolor}
\usepackage{hyperref}
\hypersetup{
    colorlinks=true,
    urlcolor=blue
}

\def\BibTeX{{\rm B\kern-.05em{\sc i\kern-.025em b}\kern-.08em
    T\kern-.1667em\lower.7ex\hbox{E}\kern-.125emX}}

\title{Stanford Pupper}
\author{\IEEEauthorblockN{Nathan Kau}
}

\title{\LARGE \bf
Stanford Pupper: A Low-Cost Agile Quadruped Robot for Benchmarking and Education}

\author{Nathan Kau$^{1}$
\thanks{$^{1}$Nathan Kau is with the Department of Mechanical Engineering,
        Stanford University, Stanford, United States
        {\tt\small fleecy@stanford.edu}}%
}

\begin{document}

\maketitle

\begin{abstract}
We present Stanford Pupper, an easily-replicated open source quadruped robot designed specifically as a benchmark platform for legged robotics research. The robot features torque-controllable brushless motors with high specific power that enable testing of impedance and torque-based machine learning and optimization control approaches. Pupper can be built from the ground up in under 8 hours for a total cost under \$2000, with all components either easily purchased or 3D printed. To rigorously compare control approaches, we introduce two benchmarks, \textit{Sprint} and \textit{Scramble} with a leaderboard maintained by Stanford Student Robotics. These benchmarks test high-speed dynamic locomotion capability, and robustness to unstructured terrain. We provide a reference controller with dynamic, omnidirectional gaits that serves as a baseline for comparison. Reproducibility is demonstrated across multiple institutions with robots made independently. All material is available at \url{https://stanfordstudentrobotics.org/quadruped-benchmark}.
\end{abstract}


\section{Introduction}
Recent successes in learning and optimization-based controllers for legged locomotion have largely been isolated to a select set of complex and sophisticated hardware platforms such as Anymal \cite{hutter2016anymal} and MIT Cheetah 3 \cite{bledt2018cheetah}. Given the complexity and slow iteration cycle of legged robots, we see little reproduction of legged robot research across different lab groups. One of the primary factors driving the complexity of these robots is that they depend on actuators that can precisely track torque commands. For learning approaches, torque control is important to ensure that the actuator models used in simulation closely parallel their physical counterparts. Similarly, optimization-based approaches often assume perfect torque control in the dynamics models, such as in the centroidal model-predictive-control method introduced in \cite{di2018dynamic}.

Additionally, new control methods are often presented without comparison to baseline or previously-developed controllers which makes it difficult to make direct comparison. More so, new methods are frequently introduced alongside new hardware, making it difficult to isolate the algorithmic contributions. This motivates the need for an open source legged robot that is 1) low-cost, 2) easily replicated, and 3) high performance with torque control.

\begin{figure}
    \centering
    \includegraphics[width=0.5\textwidth]{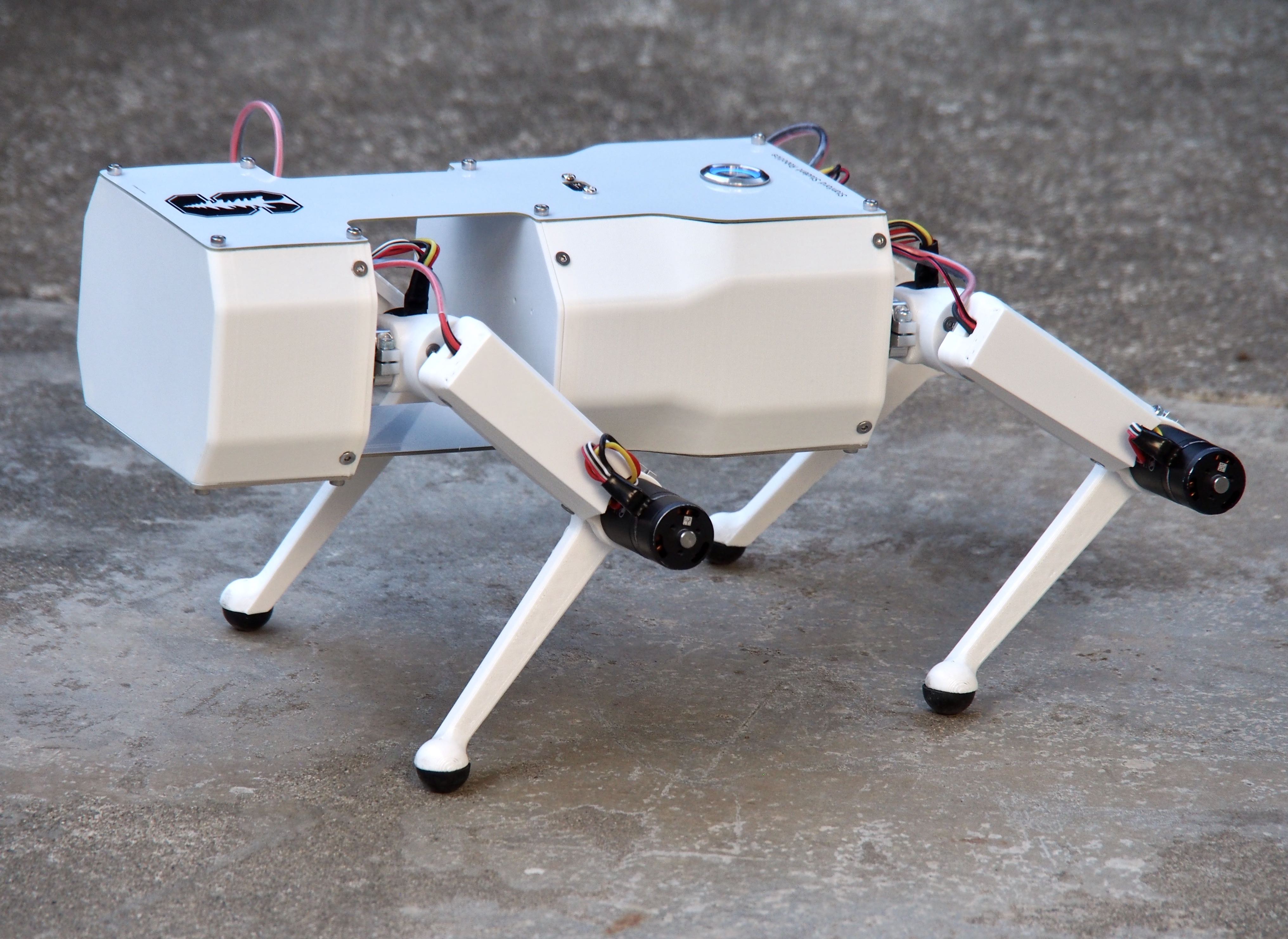}
    \caption{Stanford Pupper is an easily-replicated open source quadruped capable of dynamic locomotion.}
    \label{fig:hero}
\end{figure}

To respond to this need, we present Stanford Pupper (Fig. \ref{fig:hero}) and an associated set of benchmarks. Pupper is an open source quadruped robot designed to make legged robotics research more accessible, cost-effective, and comparable across institutions. The robot is small, lightweight, and built with off-the-shelf components. The actuators are transparent and torque-controllable, which makes the robot amenable to sim-to-real applications. Most importantly, the robot is simple and easy to build, which makes it reliable and enables fast iteration. We also present a set of benchmarks for Pupper to easily compare performance of different control algorithms across different institutions. Before detailing Pupper, we show where Pupper stands with relation to previous work.

\section{Related Work}
Previous work has explored the development of legged robots that are low-cost, easily replicated, or high performance, but to the best of our knowledge, no work has yet integrated all three characteristics into one hardware platform. At the time of this writing, two of the primary robots used for state-of-the-art learning and optimization-based control are Anymal and the MIT Cheetah variants \cite{katz2019mini}. These robots use high-bandwidth, torque-controllable actuators to accurately track control policies. However, because of their complexity and their limited availability -- Anymay is sold on a case-by-base basis, MIT Cheetah 3 is proprietary, and MIT Mini Cheetah is only partially open-source -- few researchers outside of the robots' original labs have experimented with them.

A quadruped robot that is low-cost and easily replicated is introduced by ROBEL \cite{Kumar_ROBEL}, but this platform sacrifices performance to make it accessible. They also provide three benchmarks for \textit{D'Kitty}, their quadruped robot — stand, orient, and walk — that allow rigorous comparison between learning algorithms. D’Kitty offers a low barrier to entry with a cost of \$4200, and an assembly time of less than 6 hours. The primary disadvantage of the ROBEL hardware platform is that the servo motors are too slow for agile locomotion and not transparent enough for torque control. Specifically, the servos have a maximum speed of 8 rad/s \cite{robel_motor} which is too slow for dynamic locomotion. As a reference, the Pupper actuators reach speeds up to 16 rad/s during 0.7 m/s trotting. In terms of transparency, the D’Kitty servos have a current-control mode similar to the actuators on Pupper, but D’Kitty servos have high-reduction multi-stage gear trains with high nonlinear friction which impedes accurate torque control \cite{directional}.

A handful of open source quadruped robots have attempted to reduce the gap between inexpensive and open source quadrupeds and high performance systems. Stanford Doggo \cite{kau2019stanford}, Solo \cite{grimminger2020open}, and MIT Mini Cheetah \cite{katz2019mini} use low-cost brushless drone motors and open source motor controllers to achieve record agility in certain metrics. However, all three robots required custom-machined parts and laborious assembly that ultimately make them difficult to reproduce at the scale necessary for a benchmarking platform. Of the three, Solo reduces the build complexity the most by relying more heavily on 3D printed parts. However, Solo still uses some custom-machined parts, and most importantly, uses custom actuators, transmissions, and motor controllers, which means it remains unsuitable as an accessible benchmarking platform. As we discuss below, Pupper achieves greater agility than Solo or Doggo while increasing reproducibility.

\section{Robot Overview}
Pupper is a 2.1kg, twelve degree of freedom quadruped robot capable of dynamic locomotion. We include a summary of the robot's specifications, compared against similar quadruped robots, in Tab. \ref{tab:robot-comparison}. The following subsections detail the actuator, structure, electronics, and software design of Pupper.

\begin{table}[]
\centering
\begin{tabular}{|l|l|l|l|l|}
\hline
\textbf{Robot} & \textbf{Pupper} & \textbf{Solo} & \textbf{D'Kitty} &\textbf{Mini Cheetah}\\ \hline
Mass (kg) & 2.1     & 2.2  & n/a.  & 9.   \\ \hline
Cost (USD) & 2000.  & 4700 & 4200 & n/a\textsuperscript{a}\\ \hline
Torque Control & yes  & yes  & no & yes   \\ \hline
DOF & 12 & 8 & 8 & 12 \\ \hline
\multicolumn{4}{l}{\textsuperscript{a}Estimated to be $>$10K USD including manufacturing.}
\end{tabular}
\caption{Comparison of robot specifications across low-cost quadruped robots.}
\label{tab:robot-comparison}
\end{table}

\subsection{Actuator}
The core of Stanford Pupper is its inexpensive, yet high-performance off-the-shelf actuator consisting of the M2006 brushless gearmotor \cite{m2006_store} and the C610 motor controller \cite{c610_store}.
\subsubsection{Brushless gearmotor}
\begin{figure*}[htbp]
    \centering
    \includegraphics[width=\textwidth]{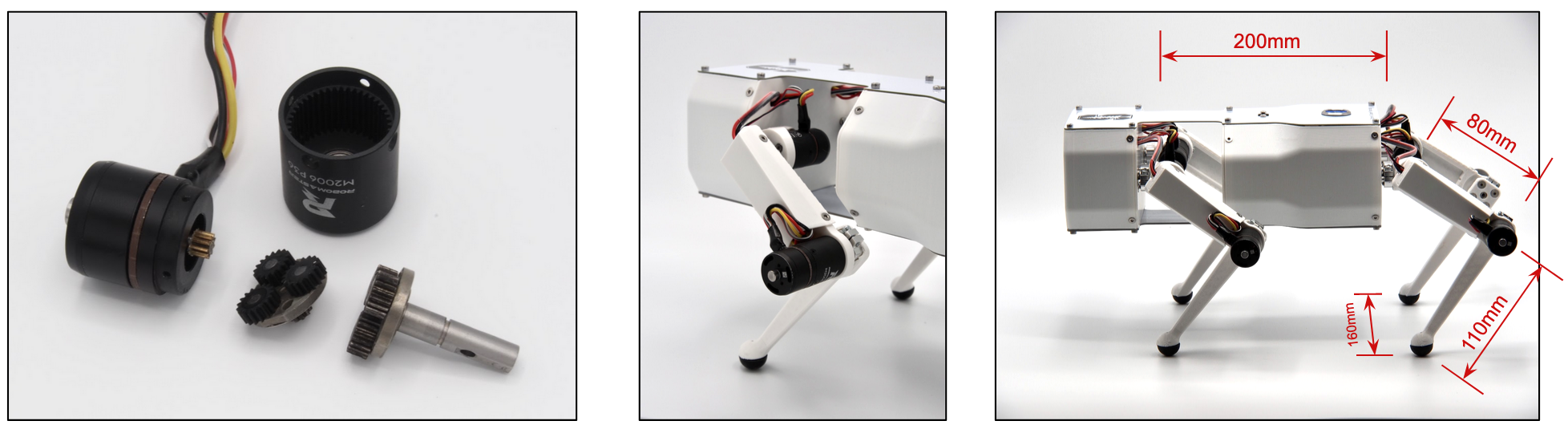}
    \caption{\textbf{(a)} Disassembled M2006 motor illustrating the BLDC, two-stage planetary reduction, and ring gear; \textbf{(b)} detailed view of the leg linkage; \textbf{(c)} schematic view of the Pupper robot indicating hip separation and link lengths.}
    \label{fig:m2006}
\end{figure*}

The M2006 brushless gearmotor is shown in Fig. \ref{fig:m2006}. The gearmotor uses a 22mm diameter brushless DC motor and integrated absolute encoder at the input followed by a 36:1 two-stage planetary reduction. The gearbox contains two stacked planetary reductions with a total ratio of 36:1. As summarized in Table \ref{tab:m2006-table}, the overall weight of the actuator is 90g and the peak torque is 1.8Nm.

\begin{figure*}[htbp]
    \centering
    \includegraphics[width=\textwidth]{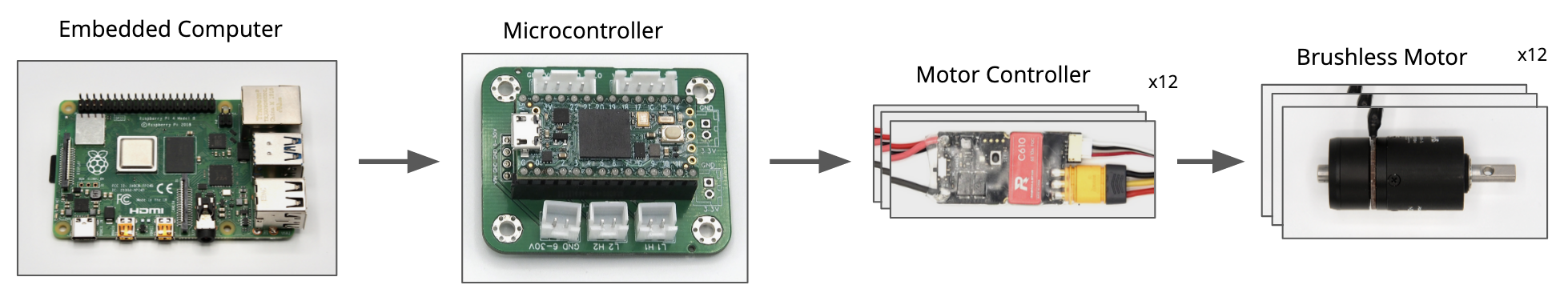}
    \caption{Electronics architecture. The embedded computer commands high-level actions to the microcontroller, which runs task- and joint-space impedance control, and commands currents to the motor controllers. The motor controllers use field-oriented-control to close the loop on motor current.}
    \label{fig:electronics}
\end{figure*}

\subsubsection{Motor controller}
The C610 motor controller uses field-oriented-control (FOC) to operate the M2006 in current-control mode. The controller supports current commands to the motor at 1kHz over a CAN connection, and reports back position, velocity, and measured current over the same CAN interface at the same rate.

\begin{table}[]
\centering
\begin{tabular}{|l|l|l|l|l|}
\hline
\textbf{Actuator} & \textbf{Pupper} & \textbf{Solo} & \textbf{D'Kitty} & \textbf{Mini Cheetah}\\ \hline
Mass (g) & 90.     & $<$150\textsuperscript{a}   & 82. & 480   \\ \hline
Maximum Speed (rad/s) & 60.  & $<$60\textsuperscript{b} & 8. &40\\ \hline
Peak Torque (Nm)   & 1.8  & 2.7  & n/a   &17  \\ \hline
Continuous Torque (Nm)  & 1.0   & n/a   & n/a   & 6.9    \\ \hline
Torque Bandwidth (Hz) & 17.     & n/a    & n/a  & 30 \\ \hline
Efficiency (\%) & 68 - 72  & n/a  & n/a   & 90 - 95\\ \hline
Output Inertia (kgm\textsuperscript{2})  & 0.0024   & n/a  & n/a & 0.0023\\ \hline
\multicolumn{5}{l}{\textsuperscript{a}Includes mass of leg segment.}\\
\multicolumn{5}{l}{\textsuperscript{b}Estimated from given specs.}
\end{tabular}
\caption{Comparison of actuator specifications between low-cost quadruped robots.}
\label{tab:m2006-table}
\end{table}

\subsection{Structure}
The robot’s frame is constructed out of 3D printed bulk heads and printed circuits boards (PCB) that double as structure and power distribution. The robot uses four identical legs, each with three actuated degrees of freedom. The feet of the robot are made of off-the-shelf wear-resistant rubber bumpers that are easily interchanged depending on the walking surface.

\subsection{Electronics}
A Teensy 4.0 microprocessor \cite{teensy} and a Raspberry Pi embedded computer \cite{raspberrypi} split responsibilities for controlling the robot. The microprocessor receives motor position, velocity, and current estimates and sends motor current commands over CAN at 1kHz. It also reads data from an inertial measurement unit (IMU) and performs onboard state estimation. The embedded computer communicates with the microprocessor over USB serial at 12mbps and is responsible for high-level motion planning, whether it be a reinforcement learning policy, model predictive control, or any other method.

\subsection{Software \& Control}
The software architecture splits the control responsibilities between the microprocessor and embedded computer. The microprocessor performs low-level actuator control, state estimation, and data logging at 1kHz. Three low-level control methods are supported. In mode 1), the microprocessor passes along motor torque commands sent by the embedded computer. In mode 2), the embedded computer sends joint angle commands and the microprocessor performs joint-space PD control to track the desired positions. This mode is intended to be used with learning-based methods, where the action space is commonly joint position commands and impedance gains. In mode 3), the embedded computer sends desired foot locations in body-relative Cartesian space and the microprocessor performs the task-space impedance control law:
\begin{equation}
\boldsymbol{F_i} = \text{Kp}(\boldsymbol{r_{ref,i}}-\boldsymbol{r_i}) + \text{Kd}(\boldsymbol{v_{ref,i}}-\boldsymbol{v_i}) + \boldsymbol{F_{ff,i}}
\end{equation}
\begin{equation}
\boldsymbol{\tau_i} = J_i^{T}\boldsymbol{F_i}
\end{equation}
where $i$ is the leg index (1 to 4), $\boldsymbol{F_i}$ is the desired foot force, Kp and Kd are impedance gains, $\boldsymbol{r_{ref,i}}$ is the reference foot position, $\boldsymbol{r_i}$ is the measured (via forward kinematics) foot position, $\boldsymbol{v_{ref,i}}$ is the reference foot velocity, $\boldsymbol{v_i}$ is the measured (via forward kinematics) foot velocity, $\boldsymbol{F_{ff,i}}$ is the feed forward force, and $J_i^T$ is the foot Jacobian.


\subsection{Robot Characterization}
\begin{figure}[htpb]
    \centering
    \includegraphics[width=0.4\textwidth]{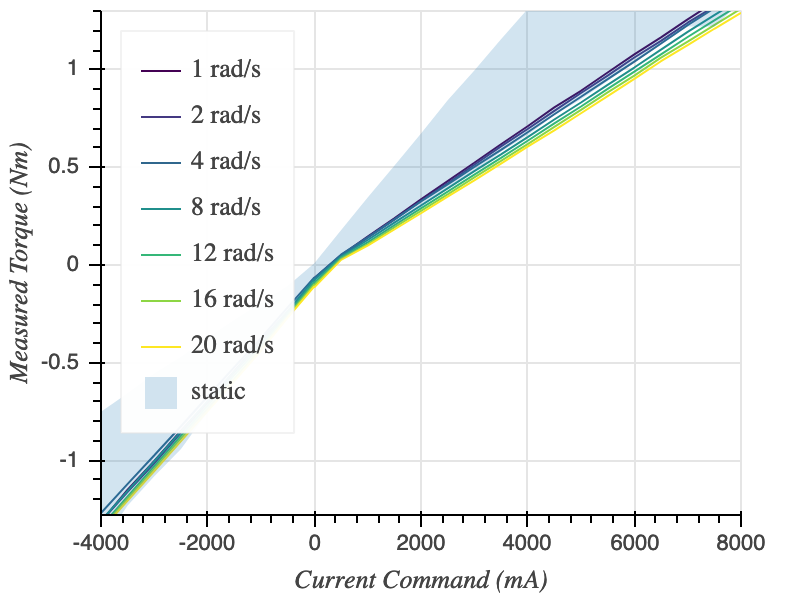}
    \caption{Measured torque versus commanded current across several motor velocities. The asymmetry between the positive and negative current cases is due to nonlinear friction in the actuator.}
    \label{fig:raw}
\end{figure}

State-of-the-art learning and optimization-based methods rely on accurate physical robot models to calculate control commands. The Pupper actuator was tested in a dynamometer to understand the actuator limits, torque-current relationship, and control bandwidth. Actuator torque was measured over varying motor speeds and motor currents to determine a best-fit surface. Fig. \ref{fig:raw} summarizes the results. The actuator's peak torque while doing positive work was 1.8Nm, while the peak torque doing negative work was 3.2Nm. The maximum continuous torque was 1.0Nm.

The motor friction was modelled well by Coulomb and damping terms, with a $R^{2}$ value over 0.999, as 

\begin{equation}
\tau_{f} = -0.021 \mathrm{Nm}\,\mathrm{sgn}(\omega) - 0.0045\frac{\mathrm{Nms}}{\mathrm{rad}}w - 10.0sgn(\omega) | \tau_{m} | \label{eq}
\end{equation}

where $\tau_f$ is the total friction in Nm, $\omega$ is the output velocity in $\frac{\mathrm{rad}}{\mathrm{s}}$, and $\tau_{m}$ is the motor torque at the gearbox input in Nm. The motor torque is given by
\begin{equation}
\tau_{m} = 0.0069 \frac{\text{Nm}}{\text{A}} i
\end{equation}
where $i$ is the motor current in A. The output torque is modeled as

\begin{equation}
    \tau_{output} = 36\tau_{m} + \tau_{f}
\end{equation}

where $\tau_{output}$ is the output torque, and the factor of $36$ comes from the 36:1 reduction ratio.

The friction model was inverted and used to predict motor current necessary to achieve desired torque commands. At constant motor velocities, arbitrary torques can be commanded within 1\% error. However, because the inverted model only corrects for velocity-dependent friction, stiction cannot be predicted, which leads to up to 28\% torque error when the actuator has zero velocity.

\begin{figure}[htpb]
    \centering
    \includegraphics[width=0.4\textwidth,height=3cm]{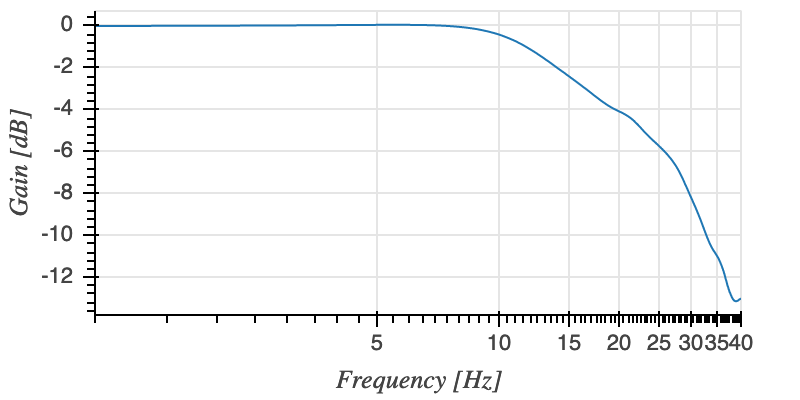}
    \caption{Actuator frequency response. A sinusoidal current command ranging from 0A to 5A was commanded while the frequency increased from 0Hz to 40Hz. The gain of the resulting torque was measured.}
    \label{fig:bode}
\end{figure}

Using the calibrated torque-current relationship, we estimated the rotor inertia by measuring output acceleration at fixed torques. The bandwidth of the actuator was determined by commanding a sinusoidal current and measuring the magnitude of the output torque as a function of frequency. A bode plot of the response is shown in Fig. \ref{fig:bode}. 

\subsection{Robot Simulator}
We offer a physically accurate Unified Robot Description Format (URDF) model for rapid experimentation and straight forward integration with robot learning environments such as OpenAI Gym \cite{brockman2016openai}. 
Work with undergraduates and high school students has highlighted a desire to integrate vision based navigation, motivating the inclusion of photo-realistic textures. The model and simulator environment can be found on the project page \cite{pupperbenchmark}.

\begin{figure}[htpb]
    \centering
    \includegraphics[width=0.47\textwidth]{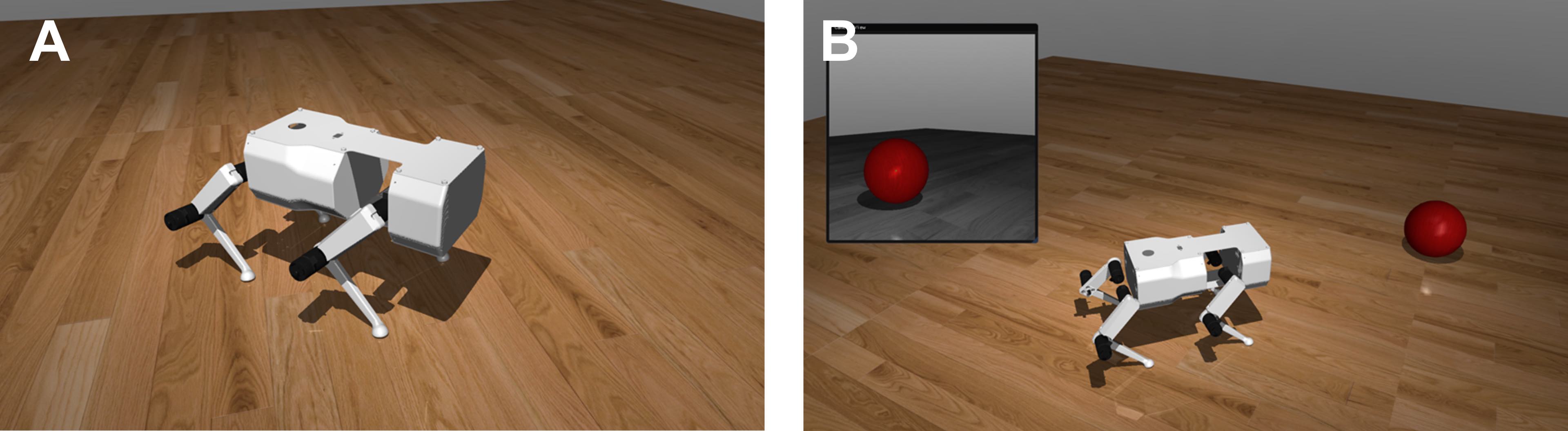}
    \caption{MuJoCo-based \cite{mujoco} simulator environment demonstrating \textbf{(a)} accurate physical model \textbf{(b)} multiple programmatically accessible camera views for closed loop visual navigation.}
    \label{fig:sim}
\end{figure}

\section{Design Discussion}
\subsection{Tradeoffs}
We optimized Pupper for easy reproducibility and experimentation while keeping performance sufficient for state-of-the-art control methods. One of the key factors that enabled easy reproducibility was reducing the robot size. While large quadruped robots have higher payload capacities and can step over larger obstacles, these advantages are not necessary to benchmark different controllers' core agility and robustness. Instead, by making Pupper small and light, we were able to use a 3D printed frame instead of custom-machine aluminum pieces. The light weight also avoids the need for costly, larger actuators. Above all, Pupper's small form factor and low weight make it easy to use in a remote-work setting since it does not need a support crane or other apparatuses for testing.

We were further able to increase accessibility while maintaining performance by using hobbyist brushless motors instead of using more expensive industrial motors. Hobbyist brushless motors with high specific torque have been used in many of the state-of-the-art quadruped robots like Mini Cheetah and Solo, but not on robots as small as Pupper. However, decreasing the motor size decreases the motor torque hyperlinearly, so larger reduction ratios are needed at small scales to maintain adequate torque. The disadvantage of higher reductions ratios are greater inertia and greater inefficiency, both qualities that hurt actuator transparency \cite{designprinciples}. While a higher quality actuator such as the Maxon ECXTQ22L BL KL A STD 24V motor and GPX22HP 35:1 gearbox would increase actuator efficiency from 72\% to 93\% and reduce inertia, the actuator cost would increase by more than ten times \cite{maxon}. In this regard Pupper sacrifices torque-controllability and transparency for greater accessibility.

An additional tradeoff made was to co-locate the knee motor at the knee joint. Unlike the common pattern of mounting the knee motor at the hip such as in Solo and MIT Mini Cheetah--which reduces leg inertia--Pupper mounts the knee motor at the knee to eliminate the need for an additional transmission. We found that the added leg inertia did not preclude agile trotting and other locomotion. 

The robot structure was simplified wherever possible for better accessibility. The 3D printed structure eliminates the need for custom-machined parts, which the authors found to be critical for helping undergraduate students take part in the work. 3D printing the majority of robots part enables experimenters to easily modify the robot geometry and material choice.

\subsection{Open Source}
The design for Pupper is entirely open source under the MIT License and all documentation are available on the project page, \url{https://stanfordstudentrobotics.org/quadruped-benchmark}. We include instructions for sourcing parts and a bill of materials, which totals less than 2000 USD for the entire robot including fabrication costs. The project page also hosts exhaustive documentation for completing the hardware assembly and software bring-up.

\section{Benchmark Tasks}
We propose a set of tasks designed for real-world benchmarking so that different controllers can be compared on equal footing across different robots and institutions. The benchmarks are designed to be repeatable and simple enough that the test can be quickly attempted for fast iteration and data collection. 
\subsection{Sprint}
\subsubsection{Task overview}
Illustrated in Fig. \ref{fig:benchmark-diagram}a, \textit{Sprint} requires Pupper to traverse an unobstructed five-meter course as fast as possible. While there are no constraints placed on the robot heading or the path it takes, Pupper must begin with zero velocity. The benchmark score is the average speed over the course, taken as the time to traverse the course divided by the five meter length. While not factored into the score, metrics including motor odometry and IMU data are logged by the onboard microcontroller to allow offline comparison of metrics including cost of transport and orientation error. Additional details are included on the project page \cite{pupperbenchmark}.
\subsubsection{Task rationale}
Achieving fast locomotion has long been a goal of the legged robotics research community, but has often not been tested under standardized conditions. High speed locomotion has garnered much research because it requires fundamental understanding of locomotion to tightly coordinate full-body motion while managing ground impacts and destabilizing perturbations. However, new control techniques are often introduced in conjunction with novel mechanical designs, which makes it difficult to disentangle the separate contributions of control and mechanical design. \textit{Sprint} is designed to provide a reproducible way to measure maximum forward speed. We hope to encourage researchers and students to compare both learned and hand designed gaits on a common platform, and to make the implementations of those gaits accessible to others.

\subsection{Scramble}
\subsubsection{Task overview}
\begin{figure}
    \centering
    \includegraphics[width=0.45\textwidth]{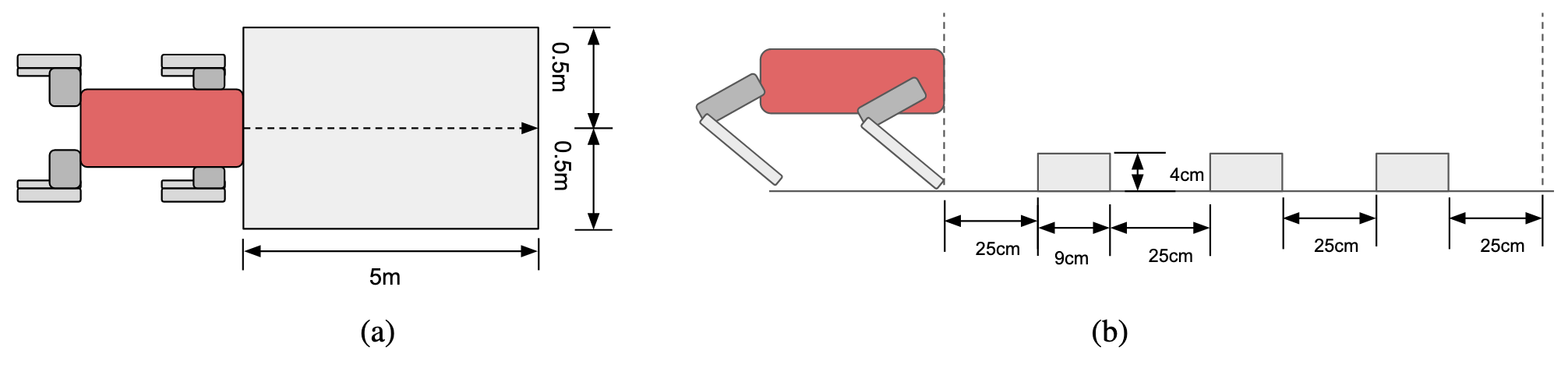}
    \caption{\textbf{(a)} The \textit{Sprint} benchmark requires the robot to travel five meters forward as fast as possible. \textbf{(b)} \textit{Scramble} tasks the robot with clambering over two tall obstacles in as little time as possible.}
    \label{fig:benchmark-diagram}
\end{figure}

This task requires Pupper to traverse a set of obstacles arranged in predefined locations shown in Fig. \ref{fig:benchmark-diagram}b. As with the \textit{Sprint} task, Pupper must start with zero velocity and the goal is to traverse the course as quickly as possible. We expect to see methods based on proprioception alone, but also methods based on vision and terrain estimation. The benchmark score is time taken to complete the course. We expect that the offline metrics may be more interesting and in some cases even more insightful than just the benchmark score. Additional details are included on the project page \cite{pupperbenchmark}.
\subsubsection{Task rationale}
This benchmark, consisting of a set of relatively tall obstacles, was designed to challenge many of the common locomotion models. In particular, the model predictive control method used in many recent quadrupeds assumes that the terrain is flat \cite{bledt2018cheetah}, \cite{di2018dynamic}. We hope this benchmark showcases generalist controllers that rely on fewer assumptions about the terrain. Like with \textit{Sprint}, we hope to see students and researchers publish their solutions for easy reproduction.

\subsection{Reference Controller}
We implemented a trotting controller to serve as a baseline for the two benchmark tasks. This controller runs on the embedded computer and generates foot position targets as a function of time and desired velocity in the horizontal plane. The architecture is similar to the position-based controller in Stanford Doggo \cite{kau2019stanford} and the Foot Trajectory Generator (FTG) architecture formalized in \cite{iscen2019policies}. Various parameters can be tuned to optimize for different gaits and terrain, including stepping height, trotting frequency, and stance/swing gait percentages. The onboard microcontroller uses task-space impedance control to calculate motor torques needed to track the foot position targets.

\begin{figure}[ht]
    \centering
    \includegraphics[width=0.45\textwidth]{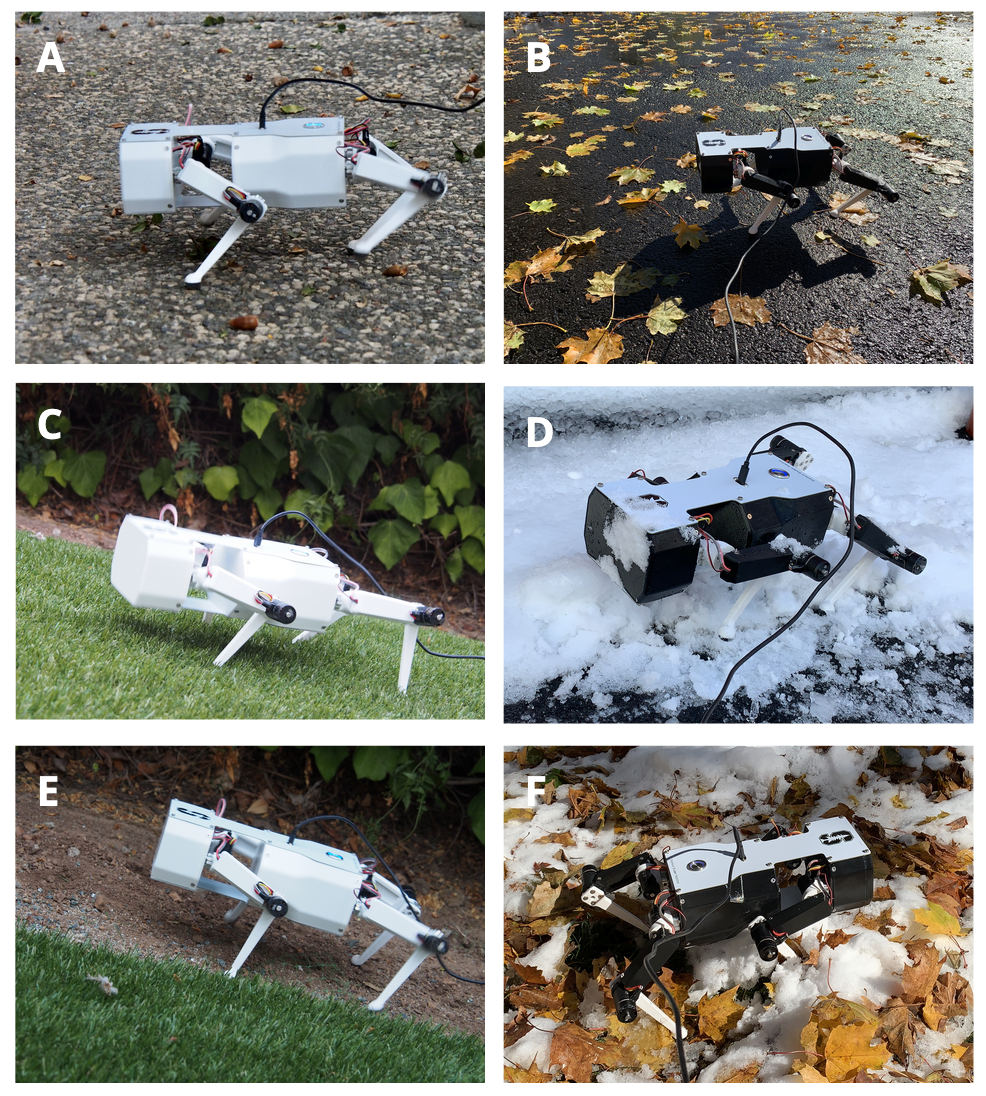}
    \caption{Pupper walking in a variety of natural environments: \textbf{(a)} pebbles, \textbf{(b)} wet ground, \textbf{(c)} grass inclined 20 degrees, \textbf{(d)} snow bank, \textbf{(e)} dirt inclined 20 degrees, \textbf{(f)} foliage}
    \label{fig:outside}
\end{figure}

\section{Experiments}
We tested the hardware platform and reference controller over a variety of terrain and demonstrated reliability, robustness and agility. Fig. \ref{fig:outside} and the supplementary video highlight the robot trotting omnidirectionally over gravel, cement, and loose natural terrain. The controller is robust to loose terrain like tanbark and pebbles, and can recover from unexpected drops going over curbs. On flat ground, the reference controller achieved a stable forward and backwards speed of 0.8 m/s, a sideways speed of 0.4 m/s, and a maximum turning rate of 2.5 rad/s. The forward speed of 0.8 m/s is comparable to the 0.8m/s trotting speed achieved with Anymal, but less than that of the more agile MIT Cheetah 3, which achieves 3.0m/s using its convex model-predictive controller.

\begin{figure}[h]
    \centering
    \includegraphics[width=0.4\textwidth, height=4cm]{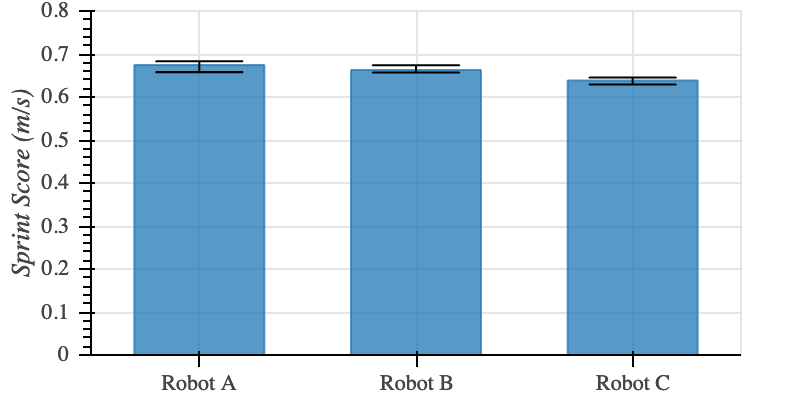}
    \caption{Comparison of benchmark scores between three different Pupper robots built at three different institutions. The stem and whiskers indicate the interquartile range of benchmark results. All three robots recorded repeatable benchmark scores within low relative error of each other.}
    \label{fig:sprint}
\end{figure}

\begin{figure}[h]
    \centering
    \includegraphics[width=0.4\textwidth, height=4cm]{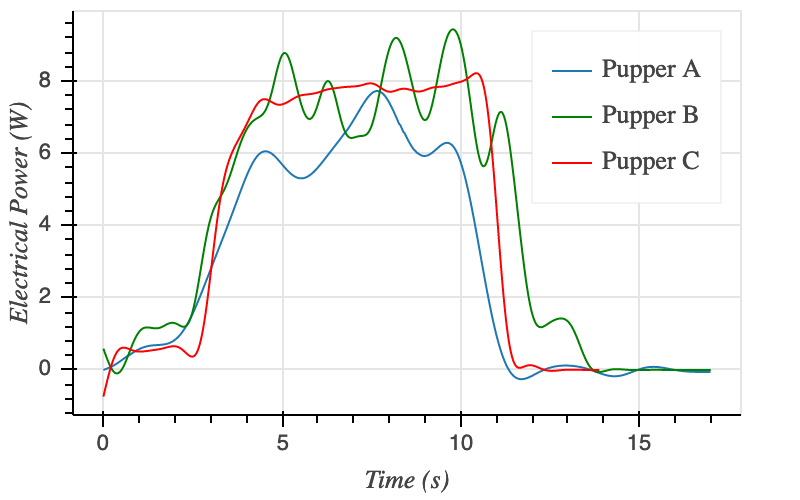}
    \caption{Comparing total motor electrical power across the three Pupper robots for the \textit{Sprint} benchmark task.}
    \label{fig:power}
\end{figure}

The benchmarks tasks were tested across three different Pupper robots, each built at different universities, in order to demonstrate replicability and to establish baseline scores. One of the robots was built by the authors, and the two others were built by undergraduate engineering students at Massachusetts Institute of Technology and Worcester Polytechnic Institute in under a day. The robots completed the \textit{Sprint} task with an average score of 0.66, and a standard deviation of 0.025. Fig. \ref{fig:sprint}, illustrates the mean and interquartile ranges of the benchmark scores across trials for each of the robots. Fig. \ref{fig:power} compares the deviation and overall consistency of electrical power used by the three robots over a single trial on the \textit{Sprint} benchmark. The \textit{Scramble} task was conducted with one robot over several trials, resulting in an average score of 34.6 with a standard deviation of 4.3.

\section{Educational Outreach}
One of the primary impacts we hope to have with this project is to introduce robotics research opportunities at the high school and college level. We have initiated several collaborations with high school and college educators to design curriculum and bring Pupper into their classrooms. Through partnership with \href{https://handsonrobotics.org}{HandsOnRobotics}, we plan to donate robots at the community and high school level to kick start a competitive robotics league centered around the Pupper platform.


\section{Conclusion}
We introduced an accessible, open source, and high performance quadruped robot to make legged robotics research more accessible and reproducible. The robot is small and lightweight which makes it ideal for when lab environments are not available such as in high school, undergraduate, and remote-work environments. Two benchmarks were introduced to provide a standard measure from which to compare progress on controller research. Through our ongoing collaborations with educators, and by the release of extensive documentation and open source design, we hope to push forward the state-of-the-art while engaging more students in legged robotics research.

\section*{Acknowledgment}
We thank the members of Stanford Student Robotics for their ongoing support of this project including Tarun Punnoose, Ian Chang, and Parthiv Krishna; Jeremy Trilling and Gregory Xie for replicating Pupper and contributing feedback; Mark Bowers for simulator integration; and Professor Zachary Manchester for discussion along the way.

\bibliographystyle{./IEEEtran}
\bibliography{./root}

\begin{thebibliography}{10}
\providecommand{\url}[1]{#1}
\csname url@samestyle\endcsname
\providecommand{\newblock}{\relax}
\providecommand{\bibinfo}[2]{#2}
\providecommand{\BIBentrySTDinterwordspacing}{\spaceskip=0pt\relax}
\providecommand{\BIBentryALTinterwordstretchfactor}{4}
\providecommand{\BIBentryALTinterwordspacing}{\spaceskip=\fontdimen2\font plus
\BIBentryALTinterwordstretchfactor\fontdimen3\font minus
  \fontdimen4\font\relax}
\providecommand{\BIBforeignlanguage}[2]{{%
\expandafter\ifx\csname l@#1\endcsname\relax
\typeout{** WARNING: IEEEtran.bst: No hyphenation pattern has been}%
\typeout{** loaded for the language `#1'. Using the pattern for}%
\typeout{** the default language instead.}%
\else
\language=\csname l@#1\endcsname
\fi
#2}}
\providecommand{\BIBdecl}{\relax}
\BIBdecl

\bibitem{hutter2016anymal}
M.~Hutter, C.~Gehring, D.~Jud, A.~Lauber, C.~D. Bellicoso, V.~Tsounis,
  J.~Hwangbo, K.~Bodie, P.~Fankhauser, M.~Bloesch \emph{et~al.}, ``Anymal-a
  highly mobile and dynamic quadrupedal robot,'' in \emph{2016 IEEE/RSJ
  International Conference on Intelligent Robots and Systems (IROS)}.\hskip 1em
  plus 0.5em minus 0.4em\relax IEEE, 2016, pp. 38--44.

\bibitem{bledt2018cheetah}
G.~Bledt, M.~J. Powell, B.~Katz, J.~Di~Carlo, P.~M. Wensing, and S.~Kim, ``Mit
  cheetah 3: Design and control of a robust, dynamic quadruped robot,'' in
  \emph{2018 IEEE/RSJ International Conference on Intelligent Robots and
  Systems (IROS)}.\hskip 1em plus 0.5em minus 0.4em\relax IEEE, 2018, pp.
  2245--2252.

\bibitem{di2018dynamic}
J.~Di~Carlo, P.~M. Wensing, B.~Katz, G.~Bledt, and S.~Kim, ``Dynamic locomotion
  in the mit cheetah 3 through convex model-predictive control,'' in \emph{2018
  IEEE/RSJ International Conference on Intelligent Robots and Systems
  (IROS)}.\hskip 1em plus 0.5em minus 0.4em\relax IEEE, 2018, pp. 1--9.

\bibitem{katz2019mini}
B.~Katz, J.~Di~Carlo, and S.~Kim, ``Mini cheetah: A platform for pushing the
  limits of dynamic quadruped control,'' in \emph{2019 International Conference
  on Robotics and Automation (ICRA)}.\hskip 1em plus 0.5em minus 0.4em\relax
  IEEE, 2019, pp. 6295--6301.

\bibitem{Kumar_ROBEL}
M.~Ahn, H.~Zhu, K.~Hartikainen, H.~Ponte, A.~Gupta, S.~Levine, and V.~Kumar,
  ``{ROBEL: RObotics BEnchmarks for Learning with low-cost robots},'' in
  \emph{Conference on Robot Learning (CoRL)}, 2019.

\bibitem{robel_motor}
``Dynamixel xm430-w210-r,''
  \url{http://www.robotis.us/dynamixel-xm430-w210-r/}, accessed: 2020-10-30.

\bibitem{directional}
A.~Wang and S.~Kim, ``Directional efficiency in geared transmissions:
  Characterization of backdrivability towards improved proprioceptive
  control,'' \emph{Proceedings - IEEE International Conference on Robotics and
  Automation}, vol. 2015, pp. 1055--1062, 06 2015.

\bibitem{kau2019stanford}
N.~Kau, A.~Schultz, N.~Ferrante, and P.~Slade, ``Stanford doggo: An
  open-source, quasi-direct-drive quadruped,'' in \emph{2019 International
  Conference on Robotics and Automation (ICRA)}.\hskip 1em plus 0.5em minus
  0.4em\relax IEEE, 2019, pp. 6309--6315.

\bibitem{grimminger2020open}
F.~Grimminger, A.~Meduri, M.~Khadiv, J.~Viereck, M.~W{\"u}thrich, M.~Naveau,
  V.~Berenz, S.~Heim, F.~Widmaier, T.~Flayols \emph{et~al.}, ``An open
  torque-controlled modular robot architecture for legged locomotion
  research,'' \emph{IEEE Robotics and Automation Letters}, vol.~5, no.~2, pp.
  3650--3657, 2020.

\bibitem{m2006_store}
``Dji m2006 brushless motor,''
  \url{https://store.dji.com/product/rm-m2006-p36-brushless-motor}, accessed:
  2020-10-30.

\bibitem{c610_store}
``Dji c610 brushless motor controller,''
  \url{https://store.dji.com/product/rm-c610-brushless-dc-motor-speed-control},
  accessed: 2020-10-30.

\bibitem{teensy}
``Teensy 4.0,'' \url{https://www.pjrc.com/store/teensy40.html}, accessed:
  2020-10-30.

\bibitem{raspberrypi}
``Raspberry pi,'' \url{https://www.raspberrypi.org/}, accessed: 2020-10-30.

\bibitem{brockman2016openai}
G.~Brockman, V.~Cheung, L.~Pettersson, J.~Schneider, J.~Schulman, J.~Tang, and
  W.~Zaremba, ``Openai gym,'' 2016.

\bibitem{pupperbenchmark}
``Stanford pupper benchmark,''
  \url{https://stanfordstudentrobotics.org/quadruped-benchmark}, accessed:
  2020-10-30.

\bibitem{mujoco}
E.~{Todorov}, T.~{Erez}, and Y.~{Tassa}, ``Mujoco: A physics engine for
  model-based control,'' in \emph{2012 IEEE/RSJ International Conference on
  Intelligent Robots and Systems}, 2012, pp. 5026--5033.

\bibitem{designprinciples}
S.~{Seok}, A.~{Wang}, {Meng Yee Chuah}, D.~{Otten}, J.~{Lang}, and S.~{Kim},
  ``Design principles for highly efficient quadrupeds and implementation on the
  mit cheetah robot,'' in \emph{2013 IEEE International Conference on Robotics
  and Automation}, 2013, pp. 3307--3312.

\bibitem{maxon}
``maxon motor,'' \url{https://www.maxongroup.com/maxon/view/content/index},
  accessed: 2020-10-30.

\bibitem{iscen2019policies}
A.~Iscen, K.~Caluwaerts, J.~Tan, T.~Zhang, E.~Coumans, V.~Sindhwani, and
  V.~Vanhoucke, ``Policies modulating trajectory generators,'' 2019.

\end{thebibliography}
\end{document}